# Triple Attention Transformer Architecture for Time-Dependent Concrete Creep Prediction

Warayut Dokduea[1], Weerachart Tangchirapat[1] and Sompote Youwai[2]*

[1]Construction Innovations and Future Infrastructures Research (CIFIR), Department of Civil Engineering, Faculty of Engineering, King Mongkut's University of Technology Thonburi, Bangkok, Thailand

[2]AI Research Group, Department of Civil Engineering, King Mongkut's University of Technology Thonburi, Bangkok, Thailand

*Corresponding author. E-mail address: sompote.you@kmutt.ac.th

## Abstract

This paper presents a novel Triple Attention Transformer Architecture for predicting time-dependent concrete creep, addressing fundamental limitations in current approaches that treat time as merely an input parameter rather than modeling the sequential nature of deformation development. By transforming concrete creep prediction into an autoregressive sequence modeling task similar to language processing, our architecture leverages the transformer's self-attention mechanisms to capture long-range dependencies in historical creep patterns. The model implements a triple-stream attention framework incorporating temporal attention for sequential progression, feature attention for material property interactions, and batch attention for inter-sample relationships. Evaluated on experimental datasets with standardized daily measurements spanning 160 days, the architecture achieves exceptional performance with mean absolute percentage error of 1.63% and $R^2$ values of 0.999 across all datasets, substantially outperforming traditional empirical models and existing machine learning approaches. Ablation studies confirm the critical role of attention mechanisms, with attention pooling contributing most significantly to model performance. SHAP analysis reveals Young's modulus as the primary predictive feature, followed by density and compressive strength, providing interpretability essential for engineering applications. A deployed web-based interface facilitates practical implementation, enabling real-time predictions using standard laboratory parameters. This work establishes the viability of applying transformer architectures to materials science problems, demonstrating the potential for data-driven approaches to revolutionize structural behavior prediction and engineering design practices.

**Keywords:** Creep, Concrete. Attention. SHAP

## 1. Introduction

Concrete, as the world's most widely used construction material, exhibits complex time-dependent deformation behaviors that significantly impact the long-term serviceability, durability, and safety of infrastructure. Compressive creep in concrete is a key time-dependent characteristic wherein deformation increases under a sustained load, even without an increment to that load (Hwang et al., 2021). The extent of this creep is also closely related to other fundamental material properties of concrete, such as its density, elastic modulus, and compressive strength (Hong et al., 2023; Wendling et al., 2018). This increase in creep deformation profoundly influences the mechanical characteristics of concrete, namely its effective modulus of elasticity over time, its capacity for stress redistribution within a structure, and its overall long-term deformability (Kammouna et al., 2019). At a microstructural level, creep in concrete is frequently associated with the propagation of fissures, especially within the interfacial transition zone between the mortar and aggregate (Giorla and Dunant, 2018). Such detrimental effects and alterations to material properties can eventually culminate in structural failures and a decline in the functional efficacy of the built construction.

However, compressive creep remains a particularly challenging phenomenon to predict accurately in concrete. Its experimental determination often requires long-duration testing, and its precise prediction is complicated by the time-dependent nature of concrete. This behavior is influenced by numerous factors, including mixture composition, environmental conditions, loading history, and aging characteristics, making reliable prediction a persistent and critical area of study in concrete technology and civil engineering (Bažant and Baweja, 1995; Gardner and Lockman, 2001). Multiple prediction models have been developed over the decades, with varying degrees of accuracy when applied to different concrete formulations and loading conditions (Fanourakis and Ballim, 2006). The complexity of this phenomenon necessitates sophisticated mathematical approaches that can account for the multifaceted interactions between material properties and external variables (Bažant and Panula, 1978; Comité Euro-International du Béton, 1993). Recent advancements in creep modeling have focused on improving prediction reliability through comprehensive databases and refined analytical techniques (Wendling et al., 2018) yet significant challenges remain in accurately forecasting long-term deformation behaviors across diverse structural applications and concrete compositions (Bažant et al., 1991; Gardner and Lockman, 2001).

Recent research on concrete creep prediction using machine learning has evolved from traditional approaches using time as a direct input feature towards more sophisticated methodologies capturing temporal dependencies. Fully connected models such as artificial neural networks (ANNs) and tree-based algorithms including Random Forest and XGBoost (Li et al., 2024; Zhu et al., 2024) have demonstrated considerable success in capturing complex non-linear relationships between material properties, environmental conditions, and creep behavior. However, these models fundamentally treat time as merely another input parameter rather than modeling the sequential nature of deformation development. (Fan et al., 2024) advanced this field by developing a Gated Recurrent Units model with Self-Attention mechanism (GSA) that explicitly models temporal progression, demonstrating superior performance for both short-term and long-term predictions by capturing the evolutionary nature of creep development through historical data patterns. Despite these improvements, a critical research gap persists in existing temporal modeling approaches. Previous studies have predominantly employed fixed time windows of 5 time steps for sequence modeling, which fundamentally constrains the model's ability to capture the comprehensive stress history governing creep behavior. This limitation is particularly problematic given that concrete creep exhibits strong dependency on the complete loading and environmental history throughout the material's service life. The inherent physics of creep deformation necessitates consideration of all previous stress states, as cumulative effects and stress history significantly influence the current deformation rate and magnitude. This concept parallels the autoregressive paradigm in large language models, where the prediction of the next token requires consideration of all preceding tokens in the sequence rather than a truncated window. Current approaches utilizing limited temporal windows fail to capture this fundamental characteristic of creep behavior, where future deformation inherently depends on the complete history of past stress states and deformation patterns. Transformer architectures, with their self-attention mechanisms capable of capturing long-range dependencies across extended temporal sequences without relying on recurrent connections or fixed window constraints, are uniquely positioned to address this limitation and enable comprehensive modeling of stress history effects in concrete creep prediction.

This paper introduces a novel Triple Attention Transformer Architecture (TATA) specifically designed for time-dependent concrete creep prediction. By leveraging the transformer's inherent ability to model sequential data through self-attention mechanisms, our approach addresses the limitations of conventional methods in capturing the complex non-linear viscoelastic behavior of concrete structures. The proposed architecture implements a triple-stream processing pathway that simultaneously captures: (1) temporal sequence dependencies that identify long-range relationships in creep deformation data, (2) material property interactions through attention mechanisms, and (3) inter-sample relationships that facilitate improved generalization across varying loading conditions (Hollmann et al., 2025; Kongkitkul et al., 2025). The feature encoding stream processes three critical concrete characteristics identified through comprehensive literature review: compressive strength, Young's modulus, and unit weight—parameters routinely obtained in standard laboratory testing protocols. These parameters were selected based on their established significance in the literature and practical accessibility in engineering applications. The architectural integration of temporal attention components, which target time-dependent patterns across creep history, and feature attention modules, which prioritize material and environmental parameters, creates a comprehensive framework designed to enhance predictive accuracy while maintaining interpretability—a critical consideration for engineering applications where model transparency directly impacts practical implementation.

This paper presents seven principal contributions to the field of concrete creep prediction through transformer-based architectures:

- Novel Transformer Architecture for Concrete Creep Prediction: A novel triple attention transformer architecture specifically designed for time-dependent concrete creep prediction, which addresses existing model limitations through a dual-stream processing pathway that simultaneously captures temporal dependencies and material-environmental interactions while incorporating specialized attention mechanisms for feature-based and batch-based encoding.

- Superior Predictive Performance: The proposed model achieves exceptional accuracy with a Mean Absolute Percentage Error (MAPE) of 1.2% per loading step, outperforming traditional models and conventional machine learning approaches, with $R^2$ values of 0.999 consistently maintained across training, validation, and test datasets, demonstrating robust performance across various concrete compositions, environmental conditions, and loading scenarios.

- Autoregressive Prediction Framework: An autoregressive approach that treats each time point as a token in a sequence, similar to language model architectures, predicting future creep values based on the complete history of measurements and effectively capturing the sequential nature of creep development through historical data patterns.

- Comprehensive Attention Mechanism Design: The architecture incorporates hybrid pooling (mean, attention, and last token pooling) for comprehensive sequence representation, multi-path feature processing (main path, feature-wise attention, and batch-level contextual modeling), and sophisticated attention masking techniques for proper sequence processing.

- Interpretability Through SHAP Analysis: The paper provides quantitative feature importance analysis showing Young's modulus as the primary predictor of concrete creep behavior (SHAP value: 0.024), with material density and compressive strength as secondary influential parameters, revealing non-linear relationship structures between material properties and creep behavior.

- Practical Implementation: The research includes a web-based application for practical industry use, comprehensive data preprocessing and normalization techniques, and Bayesian optimization for hyperparameter selection.

- Methodological Advancement: The paper demonstrates that transformer architectures are uniquely positioned to capture long-range dependencies in concrete creep, that sequential modeling approaches significantly outperform traditional empirical equations, and that the autoregressive nature of concrete creep can be effectively modeled using language model principles, thereby bridging the gap between advanced machine learning architectures and civil engineering applications to offer a practical solution for more reliable structural behavior forecasting.

The remainder of this paper is organized as follows: Section 2 provides a comprehensive review of related work in concrete creep prediction using machine learning, covering fully connected neural networks, tree-based ensemble models, sequential and convolutional architectures, and Bayesian approaches for uncertainty quantification. Section 3 presents the proposed Triple Attention Transformer Architecture (TATA), detailing its triple-stream processing framework that simultaneously captures temporal sequence dependencies, material property interactions, and inter-sample relationships through specialized attention mechanisms and autoregressive prediction methodology. Section 4 describes the data characteristics and preparation processes, including the concrete creep database compilation, temporal standardization using modified logarithmic models, and feature normalization techniques. Section 5 outlines the experimental design, encompassing model training procedures with sophisticated optimization techniques, Bayesian hyperparameter optimization, and comprehensive evaluation metrics. Section 6 presents the ablation study results, systematically quantifying the contribution of individual architectural components including attention pooling, feature attention, and batch attention mechanisms. Section 7 provides model interpretability analysis using SHAP values to identify key feature importance and reveal the predictive relationships between material properties and creep behavior. Section 8 describes the development of a web-based application deployed on Hugging Face Spaces for practical industry implementation. Finally, Section 9 discusses the findings, limitations, and implications, while Section 10 concludes with the model's achievements and future research directions for advancing transformer-based approaches in materials science applications.

## 2. Related Work

Concrete creep prediction using machine learning has evolved significantly in recent years, with researchers exploring various algorithmic approaches to model time-dependent deformation. This section reviews key developments across different machine learning architectures applied to concrete creep prediction.

### 2.1 Fully Connected Neural Network Models

Fully connected neural networks represent a fundamental architecture in concrete creep prediction, where each neuron connects to all neurons in adjacent layers. Bouras and Li (Bouras and Li, 2023) comprehensively evaluated multiple supervised machine learning algorithms for high-temperature concrete creep prediction, compiling a dataset of short-term basic creep from experimental literature. Their research identified artificial neural networks (ANN) among the most effective models, with Gaussian Process Regression (GPR) ultimately demonstrating superior performance compared to traditional empirical equations. The input variables included time, temperature, 28-day compressive strength, compressive stress, fine and coarse aggregate mass, and steel and PP fiber mass—establishing a comprehensive feature set that captures both material and environmental influences on creep behavior.

Ba Ragaa et al. (Ba Ragaa et al., 2025) advanced fully connected architectures by implementing Bayesian Neural Networks (BNN) for concrete durability assessment, which not only achieved high accuracy ($R^2$ of 0.95) but also quantified prediction uncertainty, providing more reliable assessment of degradation processes. Their innovative approach addressed the critical challenge of limited experimental data availability by expanding their dataset (comprising 403 chloride and 172 sulfate data points) using Generative Adversarial Networks (GAN). This data augmentation strategy represents a significant methodological advancement applicable to creep prediction, where experimental data collection is similarly resource-intensive and time-consuming.

Bal and Buyle-Bodin (2014) conducted pioneering work that established ANNs as viable alternatives to traditional empirical equations for concrete creep prediction. Their research demonstrated the potential of neural network approaches to capture complex material behavior without explicit mathematical formulation, setting the foundation for subsequent architectural developments in the field. Hodhod et al. (2018) developed a hybrid approach combining genetic programming with ANN to enhance predictive accuracy while maintaining model interpretability. This hybridization represents an important advancement in addressing the perceived "black box" nature of neural network models, providing engineers with greater confidence in the underlying predictive mechanisms.

## 2.2 Tree-based Ensemble Models

Tree-based algorithms have demonstrated exceptional capability in concrete creep prediction through their ensemble learning approaches and inherent interpretability. Li et al. (2024) conducted a comprehensive evaluation of Support Vector Machine (SVM), Random Forest (RF), and Extreme Gradient Boosting (XGBoost) models optimized with a Hybrid Snake Optimization Algorithm (HSOA). Their research demonstrated remarkable improvements in model accuracy, with optimized SVM, RF, and XGBoost models increasing their test set accuracies by 9.927%, 9.58%, and 14.1% respectively, with XGBoost achieving the highest precision. This work established the superiority of ensemble methods for concrete creep prediction and highlighted the critical importance of advanced optimization techniques for hyperparameter tuning.

Zhu et al (2021) further advanced tree-based approaches for Ultra-High-Performance Concrete (UHPC) by implementing extensive feature importance analysis and Bayesian optimization for hyperparameter tuning. Their optimized models achieved exceptional accuracy metrics with $R^2$ values of 0.9847, 0.9627, 0.9898, and 0.9933 for RF, ANN, XGBoost, and Light Gradient Boosting Machine (LGBM) respectively. The inherent feature ranking capabilities of tree-based models were enhanced through SHapley Additive exPlanations (SHAP), which identified loading duration, curing temperature, compressive strength at loading age, and water-to-binder ratio as the most influential parameters affecting UHPC creep behavior. This interpretability aspect represents a significant advantage for engineering applications, where understanding parameter influence is crucial for design optimization.

Gao (2025) further validated the superiority of tree-based models for specialized applications by demonstrating XGBoost's effectiveness for recycled aggregate concrete (RAC) creep prediction using a comprehensive dataset of 291 groups of in-situ and laboratory tests, optimized through random search and k-fold cross-validation methods. Their research extended ML applications to sustainable concrete formulations, addressing the increasing importance of recycled materials in construction while maintaining accurate predictive capabilities for time-dependent behavior.

Liang et al. (2022) provided additional insights into the application of ensemble machine learning models for concrete creep prediction, achieving $R^2$ values of 0.953, 0.947, and 0.946 for LGBM, XGBoost, and RF respectively on the Northwestern University database. Their comprehensive analysis of feature importance through SHAP values identified time since loading, compressive strength, loading age, relative humidity, and temperature as the most influential parameters, demonstrating consistency with theoretical understanding of concrete creep mechanisms.

Feng et al. (2022) explored multiple tree-based algorithms for recycled aggregate concrete creep prediction, establishing the efficacy of feature selection and hyperparameter optimization through grid search and k-fold cross-validation. In a complementary study, combined ensemble machine learning with SHAP analysis to identify water-cement ratio and loading age as the most significant factors influencing RAC creep behavior, providing valuable insights for sustainable concrete mix design.

## 2.3 Sequential and Convolutional Neural Network Models

Sequential models address the temporal dimension of concrete creep by incorporating time-dependent data structures capable of capturing the progressive nature of deformation. Fan et al. (2024)developed an innovative approach combining Gated Recurrent Units with a Self-Attention mechanism (GSA model) specifically for concrete creep and shrinkage prediction. Their model was trained on the Northwestern University database and validated through both short-term indoor and outdoor field tests, demonstrating superior performance compared to traditional LSTM architectures in capturing real-time outdoor temperature and humidity variations through historical data. To address the challenge of long-term prediction, Fan et al.(2024) proposed a recursive multi-step method for extrapolating short-term predictions, which was validated against long-term test data. This approach represents a significant advancement in modeling the time-dependent behavior of concrete structures under sustained loading. The sequential architecture inherently accounts for the evolutionary nature of creep development, allowing the model to learn not only from concurrent input variables but also from the historical progression of deformation. This temporal awareness enables more accurate forecasting of future behavior based on observed trends and patterns, particularly valuable for critical infrastructure requiring precise long-term performance assessment. The GSA model developed by Fan et al. (2024) demonstrated high accuracy in both short-term and long-term predictions, effectively capturing the temporal dynamics of creep progression while accounting for environmental fluctuations. This research established the potential of recurrent architectures with attention mechanisms for concrete creep prediction, setting the foundation for further innovations in sequential modeling approaches.

Zhu and Wang (2021) pioneered the application of convolutional neural networks (CNNs) for concrete creep and shrinkage prediction, overcoming challenges related to uneven data distribution on the time scale through the implementation of K-means clustering for dataset partitioning. Their research demonstrated the efficacy of CNN architectures in capturing complex spatial relationships within creep data, resulting in reliable predictions that were subsequently validated through experimental testing of reinforced concrete beams.

Cao et al. (2024) identified additional research opportunities in the domain of fly ash concrete, proposing modified B4 models for creep prediction and recovery that account for the specific properties of this sustainable concrete variant. This highlights the need for specialized models addressing the unique characteristics of emerging concrete formulations, particularly those incorporating supplementary cementitious materials or recycled aggregates.

## 2.4 Bayesian Approaches for Uncertainty Quantification

The integration of Bayesian methodologies represents a significant advancement in concrete creep prediction, providing not only accurate point estimates but also quantification of prediction uncertainty—an essential consideration for risk assessment and reliability-based design. Nguyen et al. (2025) developed a Bayesian Network framework for predicting the compressive strength of recycled aggregate concrete, demonstrating the capability of probabilistic models to capture parameter uncertainties and their propagation through the prediction process. While focused on compressive strength rather than creep specifically, this approach establishes methodological foundations applicable to creep prediction, particularly for sustainable concrete formulations incorporating recycled materials.

The Bayesian Neural Network implemented by (Ba Ragaa et al., 2025) for concrete durability assessment further exemplifies the advantage of uncertainty quantification in predictive modeling, providing confidence intervals that inform decision-making processes regarding structural assessment and maintenance planning. This probabilistic perspective represents a paradigm shift from deterministic approaches traditionally employed in concrete performance prediction, acknowledging the inherent variability in material properties, environmental conditions, and loading scenarios.

## 2.5 Research Gaps and Opportunities

The literature review demonstrates a clear evolution in concrete creep prediction methodologies, from traditional models incorporating time as a direct feature to more sophisticated sequential architectures that explicitly model temporal progression. While fully connected and tree-based models have shown considerable success in capturing the complex relationships between material properties, environmental conditions, and creep behavior, they fundamentally treat time as merely another input feature rather than modeling the sequential nature of deformation development.

A significant research gap exists in the approach to temporal modeling for concrete creep prediction. Most current models, even those categorized as sequential, do not fully leverage the historical progression of creep deformation as an explicit sequence. Although Fan et al. (2024) initially employed a sequential model architecture for creep prediction, their implementation utilized only the previous 5 time steps as input features, failing to incorporate the complete temporal history of creep behavior. This limited temporal window approach does not capture the autoregressive nature of creep behavior, where future deformation is inherently dependent on the complete history of past deformation patterns rather than discrete, truncated time segments. Furthermore, existing models inadequately address the fundamental requirement to incorporate the complete load-deformation history into the predictive framework. At a microstructural level, creep in concrete is frequently associated with the propagation of fissures, especially within the interfacial transition zone between the mortar and aggregate (Giorla and Dunant, 2018). This microstructural evolution is inherently path-dependent, where the current state of microcracking and subsequent deformation response is directly influenced by the entire loading history experienced by the material. Consequently, accurate creep prediction models must account for this cumulative damage progression by incorporating the complete load-deformation trajectory rather than relying on instantaneous or limited historical data points.

Instead of treating creep prediction as a true sequential modeling problem where each prediction builds upon the entire history of previous observations, conventional approaches typically use discrete time points as independent features. This methodology fails to account for the cumulative and memory-dependent characteristics of concrete creep, where deformation at any given time is influenced by the entire loading and environmental history experienced by the material, including the progressive microstructural changes that govern long-term mechanical behavior.

The transformer architecture, with its self-attention mechanism capable of capturing long-range dependencies across sequences, presents a promising yet unexplored approach for concrete creep prediction. By implementing an autoregressive transformer model that explicitly considers the complete historical sequence of both load application and resulting creep measurements, rather than simply using time as a model parameter or limiting input to a fixed temporal window, future research could potentially achieve significant improvements in prediction accuracy, particularly for long-term deformation forecasting where capturing temporal patterns and load history effects is crucial for reliable structural performance assessment.

## 2. Model Architecture

The proposed concrete creep prediction framework employs a multi-pathway transformer architecture that processes temporal measurement sequences and specimen-specific features through parallel computational streams, as illustrated in Fig. 1. The temporal processing pathway transforms historical creep measurements and corresponding time indicators through embedding layers, followed by positional encoding to preserve sequential information. These embedded representations are subsequently processed through an attention encoder that captures long-range temporal dependencies via self-attention mechanisms, with outputs aggregated through a hybrid pooling strategy that combines multiple representation learning approaches.

In parallel, concrete specimen features are processed through a multi-branch framework comprising three distinct computational pathways: direct feature transformation, feature-wise attention weighting, and batch-level contextual modeling. The batch attention component draws inspiration from recent advances in tabular transformer architectures (Hollmann et al., 2025), leveraging inter-sample dependencies within mini-batches to enhance statistical robustness. Theoretically, this batch-wise attention mechanism exploits cross-sample relationships to mitigate the impact of outliers and measurement noise while promoting intra-class feature consistency and improving inter-class discriminative capacity through implicit regularization effects.

The representations from both temporal and feature processing streams are subsequently integrated through a combination module, with the fused representations normalized via LayerNorm before final transformation through a feed-forward predictor network. The prediction framework implements an autoregressive architecture that forecasts the next loading step while maintaining attention over all previous loading steps, thereby generating concrete creep predictions that demonstrate robust performance across diverse specimen characteristics and loading conditions.

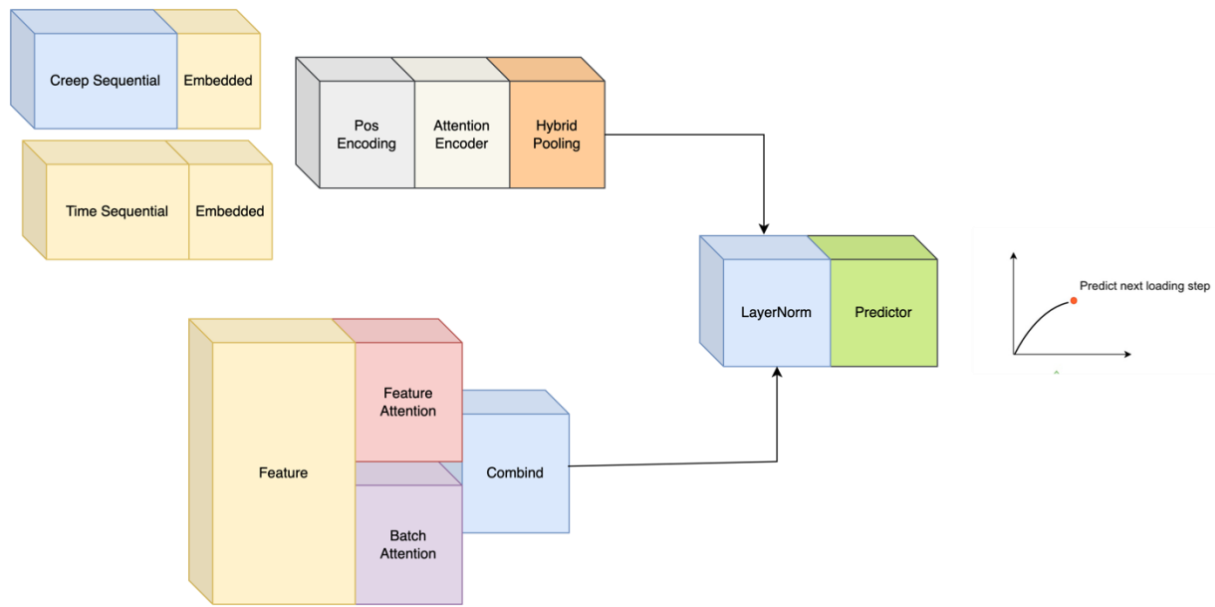

Fig. 1 The model architecture

## 2.1 Input Representation

The model processes two primary inputs:

-A temporal sequence of historical creep measurements: $X = (x_1, x_2, \ldots, x_T)$
where:
$x_t \in R$ represents a creep measurement at time $t$
$T$ is the total number of time steps in the historical sequence

-A feature vector $f \in R^F$ describing the concrete specimen's properties, where:
$F$ is the number of features.

## 2.2 Embedding Layer

Historical creep values are embedded into a higher-dimensional representation space:
$$E_{creep}(x_t) = W_{creep} \cdot x_t + b_{creep} \tag{1}$$
where:
$E_{creep}(x_t) \in R^{d_{model}}$ is the embedded representation of creep value $x_t$.
$W_{creep} \in R^{d_{model} \times 1}$ is a learnable weight matrix for creep embedding.
$b_{creep} \in R^{d_{model}}$ is a learnable bias vector for creep embedding.
$d_{model}$ is the model dimension (embedding size).

For time values $\tau = (\tau_1, \tau_2, \ldots, \tau_T)$, a parallel embedding is applied:

$$E_{time}(\tau_t) = W_{time} \cdot \tau_t + b_{time} \tag{2}$$

where:
$E_{time}(\tau_t) \in R^{d_{model}}$ is the embedded representation of time value $\tau_t$
$W_{time} \in R^{d_{model} \times 1}$ is a learnable weight matrix for time embedding
$b_{time} \in R^{d_{model}}$ is a learnable bias vector for time embedding

The combined embedding is the element-wise sum:
$$E(x_t, \tau_t) = E_{creep}(x_t) + E_{time}(\tau_t) \tag{3}$$

where $E(x_t, \tau_t) \in R^{d_{model}}$ is the combined embedding of creep and time values at position $t$

## 2.3 Positional Encoding

To preserve sequence order information, sinusoidal positional encodings are added:

$$PE_{(pos,2i)} = \sin\left(pos/10000^{2i/d_{model}}\right) PE_{(pos,2i+1)} = \cos\left(pos/10000^{2i/d_{model}}\right) \tag{4}$$

where:

$PE_{(pos,i)} \in R$ is the positional encoding value at position *pos* and dimension *i*

$pos \in 1,2,\ldots,T$ is the position in the sequence

$i \in 0,1,\ldots,d_{model} - 1$ is the dimension index.

The embedded sequence with positional encoding becomes:

$$\hat{E}(x_t, \tau_t) = E(x_t, \tau_t) + PE_t \tag{5}$$

where:

$\hat{E}(x_t, \tau_t) \in R^{d_{model}}$ is the position-aware embedding.

$PE_t \in R^{d_{model}}$ is the positional encoding vector at position *t.*

## 2.4 Feature Encoding

The feature vector undergoes multi-path processing to effectively capture various levels of interactions:

### 2.4.1 Main Path

The input vector $f \in R^{input_{dim}}$ is transformed through a two-layer feed-forward network with layer normalization and ReLU activation:

$$f_{main} = \text{LN}\Big(\text{ReLU}\big(W_2 \cdot \text{LN}\big(\text{ReLU}(W_1 \cdot f + b_1)\big) + b_2\big)\Big) \tag{6}$$

where:

$f_{main} \in R^{hidden_{dim}}$ is the main path output.

$W_1 \in R^{(hidden_{dim} \times 2) \times input_{dim}}$ and $W_2 \in R^{hidden_{dim} \times (hidden_{dim} \times 2)}$.

$b_1 \in R^{hidden_{dim} \times 2}$ and $b_2 \in R^{hidden_{dim}}$ are learnable bias vectors.

$\text{LN}(\cdot)$ represents layer normalization.

$\text{ReLU}(\cdot)$ is the rectified linear unit activation function.

$hidden_{dim}$ is the dimension of the hidden representation.

### 2.4.2 Feature-wise Projection Path

Each individual feature $f_i$ is projected to a 16-dimensional embedding and processed through multi-head attention (4 heads):

$$f_i' = \text{Projection}(f_i) \in R^{16} \tag{7}$$

$$f_{feat} = \text{LN}\big(\text{MHA}_{feat}(f', f', f') + f'\big) \tag{8}$$

where:

$f_i \in R$ , $f_i$ is the *i*-th feature value

$f'_i \in R^{16}$ is the projected representation of feature $f_i$

$f' \in R^{input_{dim} \times 16}$ is the collection of all projected feature representations.

$\text{MHA}_{\text{feat}}(\cdot)$ is multi-head attention with 4 heads for feature-wise processing.

$f_{feat} \in R^{input_{dim} \times 16}$ is the feature-wise attention output.

The feature-wise attention output is flattened to a vector of dimension $(input_{dim} \times 16)$ .

### 2.4.3 Batch Attention Path

The input vector is projected to a fixed embedding dimension and processed through multi-head attention:

$$f_{proj} = \text{Projection}batch(f) \in R^{16} \tag{9}$$

$$fbatch = \text{LN}\big(\text{MHA}batch(fproj, f_{proj}, f_{proj}) + f_{proj}\big) \tag{10}$$

where:

$f_{proj} \in R^{16}$ is the batch-projected representation $f$.

$\text{MHA}_{batch}(\cdot)$ is multi-head attention with 4 heads for batch processing.

$f_{batch} \in R^{16}\$$ is the batch attention output,

### 2.4.4 Integration

The outputs from all three paths are concatenated and integrated:

$$f_{combined} = [f_{main}; f_{feat}; f_{batch}] \tag{11}$$

$$f_{output} = \text{ReLU}\Big(\text{LN}\big(W_{integration} \cdot f_{combined} + b_{integration}\big)\Big) \tag{12}$$

where:

$f_{combined} \in R^{(hidden_{dim} + input_{dim} \times 16 + 16)}$ is the concatenated vector of all pathway outputs.

$W_{integration} \in R^{hidden_{dim} \times (hidden_{dim} + input_{dim} \times 16 + 16)}$ is a learnable weight matrix.

$b_{integration} \in R^{hidden_{dim}}\$$ is a learnable bias vector.

$f_{output} \in R^{hidden_{dim}}$ is the final encoded feature representation.

This multi-path architecture enables the model to capture different levels of feature interactions: direct transformations, feature-to-feature relationships, and batch-level contextual information.

### 2.5 Self-Attention Mechanism

For the embedded sequence $\hat{E} \in R^{T \times d_{model}}$, multi-head self-attention is computed as:

$$\text{MultiHead}(\hat{E}) = \text{Concat}(head_1, \ldots, head_h)W^O \tag{13}$$

where:

$\text{MultiHead}(\hat{E}) \in R^{T \times d_{model}}$ is the multi-head attention output.

$head_i \in R^{T \times (d_{model}/h)}$ is the output of the $i$-th attention head.

$h$ is the number of attention heads.

$W^O \in R^{d_{model} \times d_{model}}$ is a learnable output projection matrix.

$\text{Concat}(\cdot)$ represents the concatenation operation along the feature dimension.

Each attention head is calculated as:

$$head_i = \text{Attention}(\hat{E}W_i^Q, \hat{E}W_i^K, \hat{E}W_i^V) \tag{14}$$

where:

$W_i^Q \in R^{d_{model} \times d_k}$ is the query projection matrix for head $i$.

$W_i^K \in R^{d_{model} \times d_k}$ is the key projection matrix for head $i$.

$W_i^V \in R^{d_{model} \times d_v}$ is the value projection matrix for head $i$.

$d_k = d_v = d_{model}/h$ are the dimensions of keys and values in each head.

The scaled dot-product attention is defined as:

$$\text{Attention}(Q, K, V) = \text{softmax}\left(\frac{QK^T}{\sqrt{d_k}}\right)V \tag{15}$$

where:

$Q \in R^{T \times d_k}$ is the query matrix.

$K \in R^{T \times d_k}$ is the key matrix.

$V \in R^{T \times d_v}$ is the value matrix

$\text{softmax}(\cdot)$ is applied row-wise.

## 2.6 Feed-Forward Network

Each position in the sequence is processed by a position-wise feed-forward network:

$$\text{FFN}(x) = \text{LN}(x + \text{Dropout}(W_2 \cdot \text{ReLU}(W_1 \cdot x + b_1) + b_2)) \tag{16}$$

where:

$x \in R^{d_{model}}$ is the input vector at a specific position.

$W_1 \in R^{d_{ff} \times d_{model}}$ and $W_2 \in R^{d_{model} \times d_{ff}}$ are learnable weight matrices.

$b_1 \in R^{d_{ff}}$ and $b_2 \in R^{d_{model}}$ are learnable bias vectors.

$d_{ff}$ is the feed-forward network's inner dimension.

$\text{Dropout}(\cdot)$ is the dropout regularization function.

$\text{LN}(\cdot)$ is the layer normalization.

## 2.7 Encoder Layer

A complete encoder layer combines self-attention and feed-forward networks:

$$x' = \text{SelfAttention}(x, \text{mask}) \tag{17}$$

$$\text{EncoderLayer}(x) = \text{FFN}(x') \tag{18}$$

where:

$x \in R^{T \times d_{model}}$ is the input sequence representation.
$\text{mask} \in R^{T \times T}$ is an optional attention mask to prevent attending to padding tokens.
$x' \in R^{T \times d_{model}}$ is the intermediate representation after self-attention.
$\text{EncoderLayer}(x) \in R^{T \times d_{model}}$ is the output of the encoder layer.

The model stacks $N$ such encoder layers, where each layer progressively refines the sequence representation. The final output $\text{Encoder}N(\hat{E}) \in R^{T \times dmodel}$ contains contextual information about the creep history sequence.

## 2.8 Context Integration

The model integrates sequence representation with feature information through a hybrid pooling approach:

### 2.8.1 Hybrid Context Vector Extraction

This approach combines three complementary pooling strategies to extract comprehensive representations:

**Mean Pooling:** Captures the overall trend by averaging all valid token representations:

$$c_{\text{mean}} = \frac{\sum_{t=1}^{T} \text{Encoder}N(\hat{E})i,t \cdot m_{i,t}}{\sum_{t=1}^{T} m_{i,t}} \tag{19}$$

where:
$c_{\text{mean}} \in R^{d_{model}}$ is the mean-pooled context vector.
$\text{Encoder}N(\hat{E})i, t \in R^{d_{model}}$ is the encoded representation at position $t$ for sequence $i$.
$m_{i,t} \in 0,1$ is a mask value indicating whether position $t$ is valid (1) or padding (0).

**Attention Pooling:** Identifies and emphasizes critical points in the sequence:

$$c_{\text{attn}} = \sum_{t=1}^{T} \alpha_t \cdot \text{Encoder}N(\hat{E})i,t \tag{20}$$

where attention weights $\alpha_t$ are computed as:

$$\alpha_t = \frac{\exp(W_{\text{attn}} \cdot \text{Encoder}N(\hat{E})i,t)}{\sum_{s=1}^{T} \exp(W_{\text{attn}} \cdot \text{Encoder}N(\hat{E})i,s) \cdot m_{i,s}} \tag{21}$$

where:
$c_{\text{attn}} \in R^{d_{model}}$ is the attention-pooled context vector.

$\alpha_t \in R$ is the attention weight for position $t$.

$W_{attn} \in R^{1 \times d_{model}}$ is a learnable parameter vector for attention scoring.

**Last Token Pooling:** Preserves the most recent measurement:

$$c_{last} = \text{Encoder}N(\hat{E})i, l_i \tag{22}$$

where:

$c_{last} \in R^{d_{model}}$ is the last-token context vector.

$l_i$ is the index of the last valid token for sequence $i$.

**Integration of Pooling Methods:**

$$c_{hybrid} = \tanh\left(W_{hyb}[c_{mean} \oplus c_{attn} \oplus c_{last}]\right) \tag{23}$$

where:

$c_{hybrid} \in R^{d_{model}}$ is the hybrid context vector.

$\oplus$ denotes concatenation.

$W_{hyb} \in R^{d_{model} \times 3d_{model}}\$$ is a learnable weight matrix.

## 2.8.2 Feature Integration

The context vector is integrated with the encoded feature vector:

$$h = \tanh\left(W_{comb} \cdot [c_{hybrid} \oplus f_{encoded}]\right) \tag{24}$$

where:

$W_{comb} \in R^{d_{model} \times 2d_{model}}$ is a learnable weight matrix.

## 2.8.3 Normalization

The integrated representation undergoes layer normalization:

$$\text{LN}(h) = \gamma \odot \frac{h - \mu_h}{\sqrt{\sigma_h^2 + \epsilon}} + \beta \tag{25}$$

where:

$\mu_h \in R$ is the mean of $h$ computed across the feature dimension

$\sigma_h^2 \in R$ is the variance of $h$ computed across the feature dimension.

$\gamma, \beta \in R^{d_{model}}$ are learnable scale and shift parameters.

$\epsilon$ is a small constant (typically $10^{-5}$ or $10^{-6}$) for numerical stability.

$\odot$ denotes element-wise multiplication.

## 2.9 Prediction Layer

The final prediction is generated through a two-layer feedforward neural network:

$$h_{intermediate} = \text{ReLU}\left(\text{Dropout}\left(W_{pred1} \cdot h + b_{pred1}\right)\right) \tag{26}$$

$$\hat{y} = W_{pred2} \cdot h_{intermediate} + b_{pred2} \tag{27}$$





where:

$h \in R^{d_{model}}$ is the normalized integrated representation.

$W_{pred1} \in R^{d_{intermediate} \times d_{model}}$ is the first layer weight matrix.

$b_{pred1} \in R^{d_{intermediate}}$ is the first layer bias vector.

$W_{pred2} \in R^{target_{len} \times d_{intermediate}}$ is the second layer weight matrix.

$b_{pred2} \in R^{target_{len}}$ is the second layer bias vector.

$\hat{y} \in R^{target_{len}}$ represents predicted future creep values.

The architecture incorporates N sequential encoder layers, with each layer iteratively refining the sequence representation. This hierarchical processing enables progressive abstraction and feature extraction from the input data. The terminal layer output, $\hat{y} \in R^{target_{len}}$, encapsulates comprehensive contextual information regarding the concrete creep history sequence. For the purposes of this investigation, the target prediction is constrained to a single dimension, specifically forecasting the subsequent temporal increment of concrete creep deformation.

## 3. Data Characteristics

The concrete compressive creep database was obtained from the Concrete and Materials Testing Laboratory at King Mongkut's University of Technology Thonburi, Thailand. This database comprises 42 creep test curves from tests conducted in accordance with ASTM C512/C512M (ASTM International, 2015), along with an additional 25 data sourced from other journal articles (Choi et al., 2025; Gao et al., 2025; Hong et al., 2024; Rossi et al., 2013; Zhao et al., 2016). The compilation of this comprehensive dataset was necessary to establish a robust foundation for creep behavior analysis, as the variability in concrete creep response requires extensive experimental data to capture the full range of material behavior under different conditions. The inclusion of data from multiple sources ensures representativeness across various concrete mix designs, curing conditions, and loading scenarios, thereby enhancing the statistical significance and generalizability of subsequent analytical investigations.

A comprehensive investigation of time-dependent deformation behavior in concrete was undertaken, with particular emphasis on compressive creep strain characterization. Creep strain measurements were performed on cylindrical specimens in accordance with ASTM C512/C512M (2015) standardized procedures. Upon completion of the prescribed curing period, all test specimens were transferred to a controlled environmental chamber maintaining ambient conditions of 50 ± 4% relative humidity and 23 ± 2°C temperature. The experimental setup for creep strain determination (Fig. 2) comprised a hydraulic spring-loaded testing frame wherein specimens were systematically positioned. Initial specimen length measurements were obtained through the installation of two datum disks per specimen, providing reference points for subsequent deformation monitoring. A sustained compressive load equivalent to 40% of the material's compressive strength at the loading age was applied via hydraulic jack mechanism and maintained throughout the testing duration. Strain data acquisition followed a predetermined schedule consisting of immediate post-loading measurements, followed by daily monitoring during the





initial seven-day period, and weekly measurements thereafter until asymptotic creep strain values were achieved. This monitoring protocol ensured comprehensive characterization of both primary and secondary creep phases, enabling accurate determination of long-term deformation behavior under sustained loading conditions.

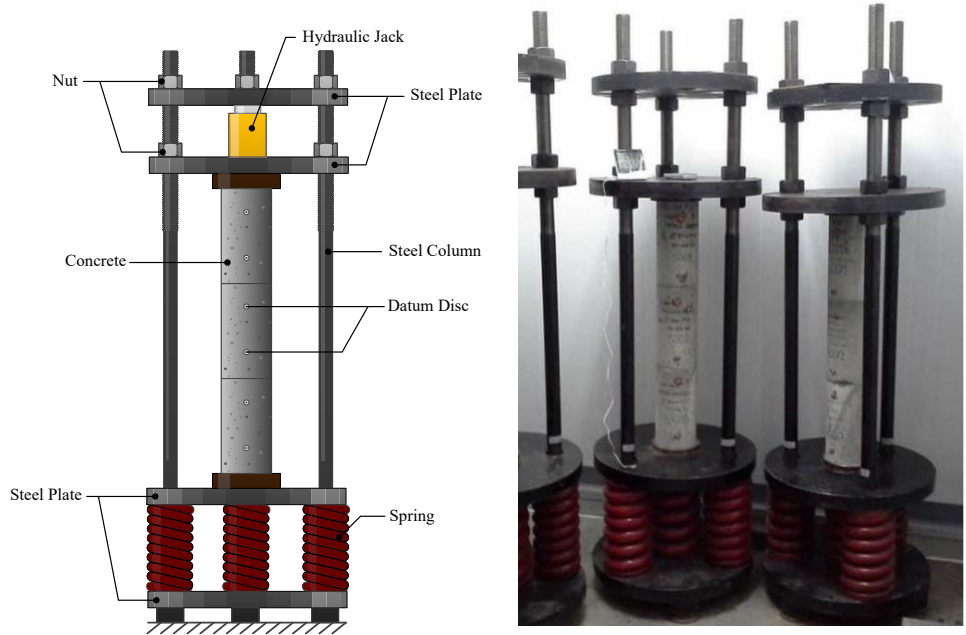

**Fig. 2** Creep test setups

The experimental concrete creep data exhibited significant methodological challenges due to non-uniform temporal sampling across specimens. These irregular temporal intervals, resulting from practical testing schedule limitations and variable measurement protocols, necessitated a standardization approach for comprehensive comparative analysis. To address this heterogeneity, a systematic data regularization methodology was implemented. The standardization procedure employed a modified logarithmic model:

$$\text{Creep} = a \cdot (1 - \exp(-b \cdot t^c)) \tag{28}$$

where $a$, $b$ and $c$ are model parameters optimized for each specimen, $t$ represents time, and $exp$ denotes the exponential function.

This model was selected for its capacity to capture the characteristic non-linear viscoelastic response of cementitious materials while maintaining physical consistency with zero initial creep. Parameter optimization was conducted utilizing the Levenberg-Marquardt algorithm for non-linear least squares regression, with a maximum function evaluation threshold of 10,000 iterations to ensure robust convergence. For each specimen, the optimized model parameters ($a$, $b$, $c$) were determined through minimization of the residual sum of squares between experimental measurements and theoretical predictions. These calibrated parameters were subsequently utilized to generate uniformly spaced daily creep predictions extending to 160 days. The curve fitting procedure yielded an $R^2$ value of 0.9950. The interpolation was configured to produce time steps of 1 day for each sample, as illustrated in Fig. 3. This approach not only standardized the temporal





domain for direct inter-specimen comparison but also established a consistent dataset structure conducive to implementation in sequential machine learning models, thereby facilitating more advanced predictive applications and pattern recognition in time-dependent concrete behavior.

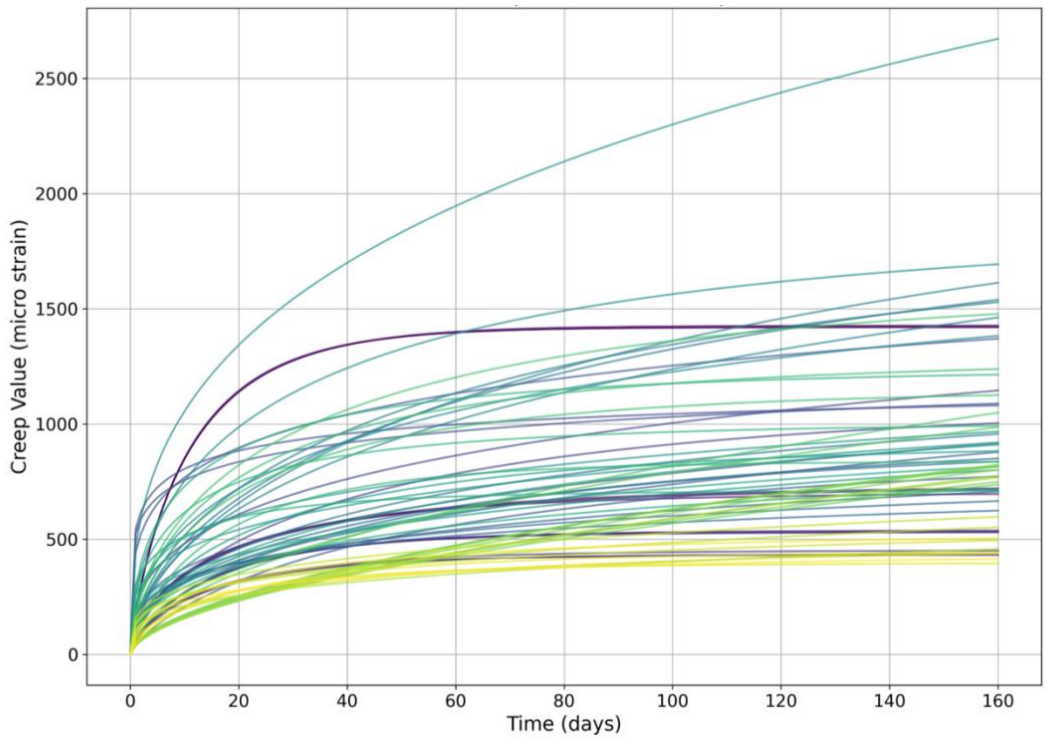

Fig. 3 The Creep deformation value with time of concrete





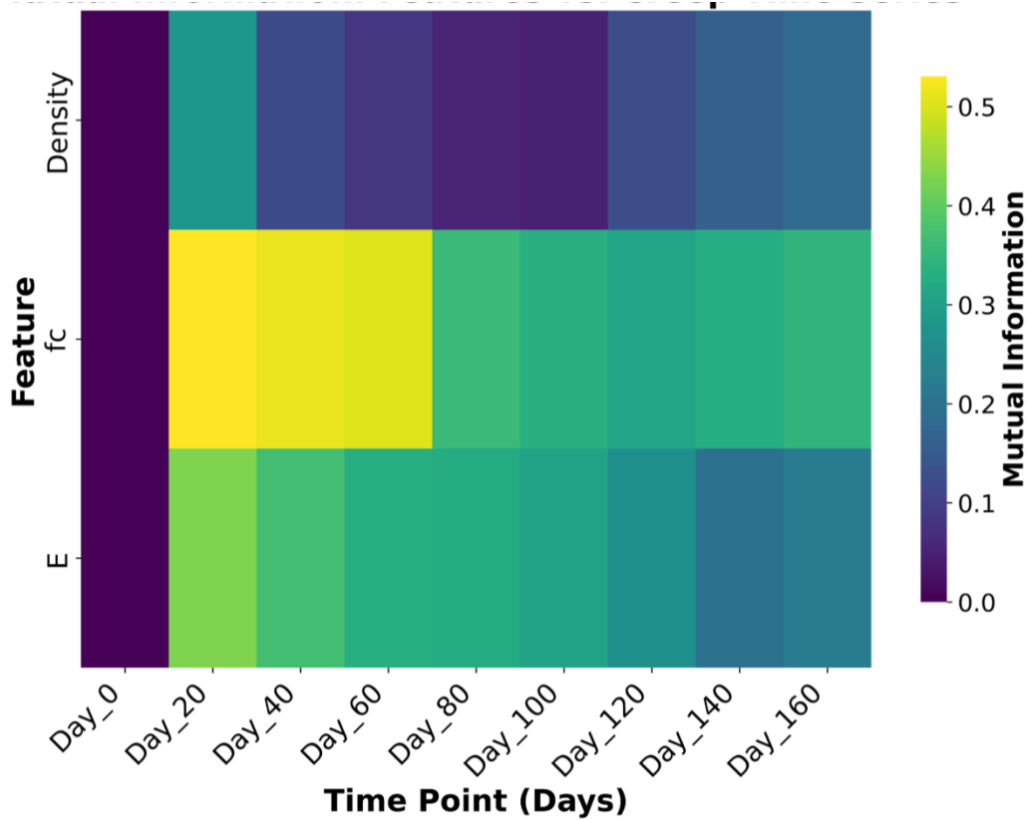

Fig. 4. Quantification of feature relevance via mutual information analysis with respect to concrete creep strain

The quantification of predictive relationships between concrete material properties and time-dependent creep strain was systematically evaluated through mutual information analysis, as illustrated in Fig. 4. The investigated features—compressive strength (fc), elastic modulus (E), and density—represent standard parameters obtained through conventional laboratory testing protocols. Analysis was conducted on a dataset comprising 65 concrete creep specimens. The mutual information heatmap reveals distinctive temporal-feature dependencies, with compressive strength demonstrating substantially elevated information content (0.4-0.5) during early to intermediate creep periods (Day 20-60), indicating its pronounced predictive capacity during this critical phase of viscoelastic deformation. Elastic modulus exhibited moderate yet consistent mutual information values (approximately 0.3-0.4) across the temporal domain, suggesting persistent relevance throughout the creep development process. In contrast, density displayed markedly lower mutual information coefficients (0.0-0.2), indicating comparatively limited predictive utility for creep strain estimation. A notable temporal pattern was observed wherein the information content of all material properties exhibited progressive attenuation at extended time points (Day 100-160), suggesting diminished predictive capacity as long-term creep mechanisms become increasingly dominant. These findings provide quantitative substantiation for feature prioritization in predictive modeling frameworks, with compressive strength and elastic modulus containing significantly more information about concrete creep behavior than density, particularly during the formative stages of viscoelastic deformation development.





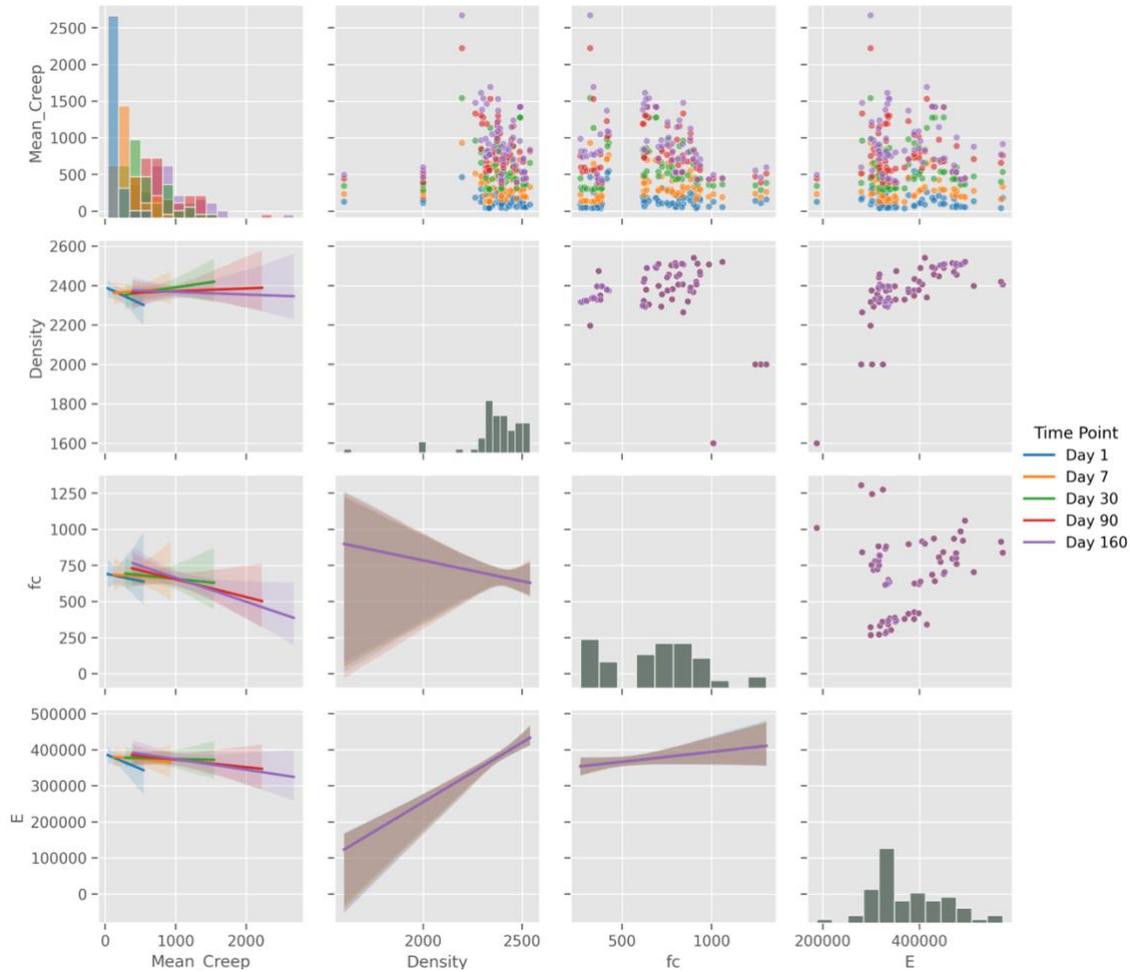

Fig. 5. Temporal distribution of concrete creep strain values and corresponding material property correlations across test duration

The scatter plot matrix presented in Fig. 5 provides a comprehensive multivariate analysis of concrete creep strain relationships with fundamental material properties across five temporal reference points (Day 1, 7, 30, 90, and 160). Examination of the temporal distribution reveals pronounced viscoelastic evolution characterized by progressively increasing mean creep strain values and heightened data dispersion at extended time intervals. The frequency distribution of creep strain demonstrates notable positive skewness with modal concentration in the lower strain region (0-500 με) and decreasing occurrence frequency at elevated strain magnitudes. Analysis of material property correlations reveals several statistically significant relationships: density exhibits a moderate positive correlation with creep strain, while both compressive strength (fc) and elastic modulus (E) demonstrate inverse relationships with strain development, consistent with established viscoelastic mechanisms wherein enhanced material stiffness and strength contribute to improved creep resistance. The temporal dimension introduces additional complexity, with regression lines across different time points maintaining directional consistency while exhibiting systematic vertical displacement, indicating preservation of fundamental structure-property relationships despite absolute strain magnitude increases. This temporal stability suggests that initial material characterization parameters maintain predictive utility throughout the creep





development process. Notable temporal effects include: (1) increasing scatter and heteroscedasticity at later time points, suggesting emergence of secondary mechanisms not fully captured by initial material properties; (2) convergence of regression confidence intervals at Day 90 and Day 160 for the E-creep relationship, indicating potential stabilization of elastic modulus influence at later stages; and (3) differential temporal sensitivity across material properties, with density correlation slopes showing greater time-dependent variation than compressive strength correlations. These temporal patterns suggest a mechanistic transition from early-stage creep behavior dominated by initial elastic response to later-stage behavior increasingly influenced by long-term microstructural rearrangements and potential redistribution of internal stresses within the cementitious matrix. The preservation of correlation directionality despite this mechanistic evolution provides theoretical support for time-dependent predictive models incorporating these fundamental material parameters.

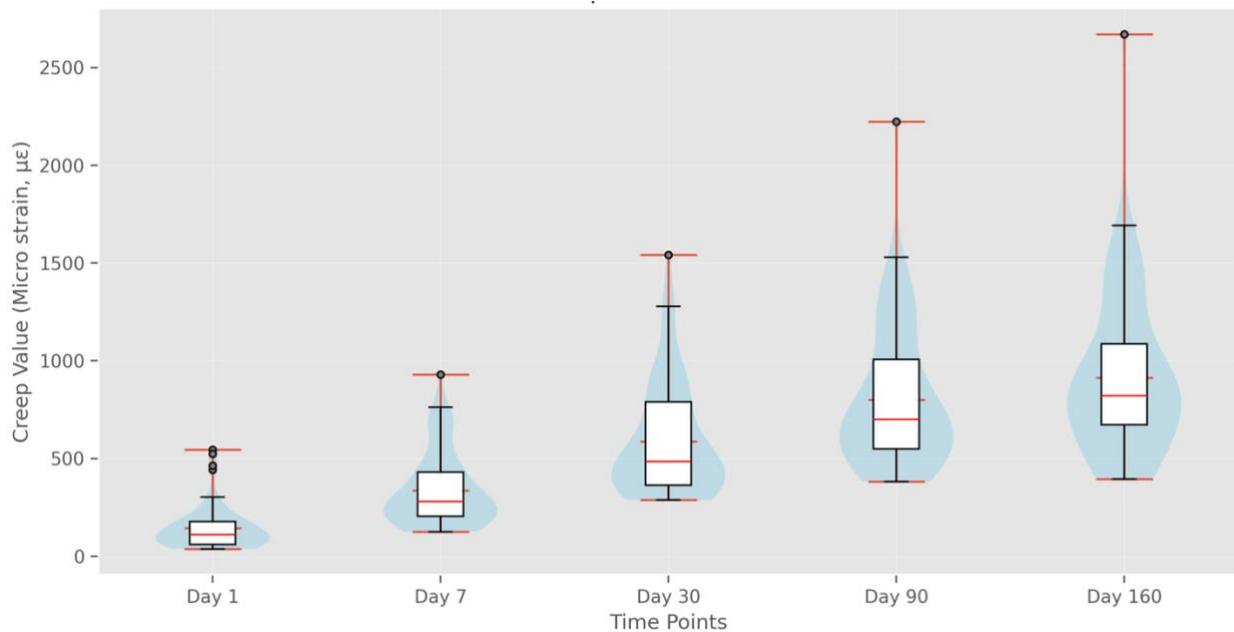

Fig. 6. Temporal evolution of concrete creep strain distribution across critical test intervals

The violin-box plot in Fig. 6 demonstrates the temporal evolution of concrete creep strain across five measurement intervals (Days 1-160). Statistical analysis reveals a systematic increase in both median values (from ~100 με to ~800 με) and distribution variance, with emergent bimodality at later stages. The distribution morphology transitions from a narrow, positively skewed curve at Day 1 to broader, more symmetric patterns with bimodal characteristics by Days 90-160, indicating divergent long-term behavior patterns. This heterogeneous creep response can be attributed to the differential influence of concrete material properties—compressive strength, elastic modulus, and density—on long-term viscoelastic behavior. The widening interquartile ranges (approximately five-fold increase from Day 1 to Day 160) and elongated upper whiskers (maximum values increasing from 550 με to 2700 με) reflect how initial material variations amplify over time under sustained loading. Specimens with higher compressive strength and elastic modulus generally populate the lower portions of each distribution, demonstrating enhanced creep resistance, while specimens with lower mechanical properties contribute to the





upper extremes of the distributions, experiencing accelerated strain development. The pronounced upper tail elongation at later time points indicates that certain material compositions undergo disproportionate strain amplification during extended loading durations, potentially due to increased microcracking, progressive C-S-H restructuring, or enhanced moisture migration within more susceptible cementitious matrices. This material-dependent divergence explains both the observed expansion in statistical dispersion and the development of distinct response clusters in the long-term creep distribution.

## 4. Experiments

### 4.1 Data Preparation

The model employs an LLM-style framework to process concrete creep measurements with a fixed structure of 160 data points, representing 160 days of observations for each specimen. Our methodology organizes each concrete specimen's data as a structured collection of historical creep measurements, corresponding time points, material property vectors, and target future creep values. Although all specimens have the same sequence length, the LLM-style approach still provides significant advantages in capturing the temporal patterns and dependencies across the full 160-day period. The model uses an autoregressive approach similar to Large Language Models (LLMs), treating each of the 160 time points as tokens in a sequence. Just as an LLM predicts the next token based on all previous tokens, our model predicts future creep values based on the entire history of measurements up to the prediction point. This parallels how language models process text sequences, allowing the model to develop a rich understanding of how past creep behavior influences future development within the consistent 160-day timeframe.

Our implementation incorporates sophisticated attention masking techniques to ensure proper sequence processing. The attention mask is constructed as a 3D tensor that guides the self-attention mechanism across the entire sequence. This attention mask $M_{attn}$ is formulated as:

$$M_{attn}[i, j, k] = \begin{Bmatrix} 1, \text{if position } k \text{ is invalid for batch } i \\ 0, otherwise \end{Bmatrix} \tag{29}$$

where $i$ is the batch index, while $j$ and $k$ represent the query and key positions in the self-attention mechanism. This attention mask is applied in the self-attention computation:

$$\text{Attention}(Q, K, V) = \text{softmax}\left( \frac{Q K^T}{\sqrt{d_k}} + M_{attn} \cdot (-\infty) \right) V \tag{30}$$

By multiplying masked positions by negative infinity before the softmax operation, the attention scores for these positions effectively become zero, ensuring that each position in the sequence only attends to valid positions. This creates a precise pattern of information flow that enables the model to capture the progressive nature of concrete creep development across the 160-day period.

Effective normalization techniques are implemented to stabilize the training process. Creep values are scaled down by a factor of 1000 to bring them into a numerically stable range. Time points undergo logarithmic transformation to compress the time scale while preserving relative ordering, enhancing the model's capacity to learn temporal patterns across different phases of the 160-day period. Material features are standardized using z-score normalization to ensure all





features contribute proportionally regardless of their original scales. Proper normalization is essential for stabilizing the training process:

**Creep Value Normalization**: We apply a scaling factor to standardize the creep values, addressing the typically large magnitude differences:

$$\widetilde{x}_t = \frac{x_t}{\alpha} \tag{31}$$

Where $\alpha$ (typically 1000 in our implementation) to bring creep values into a numerically stable range.

**Time Value Normalization**: Time points often span multiple orders of magnitude and exhibit non-linear progression. We apply logarithmic transformation:

$$\widetilde{\tau}_t = \ln(1 + \tau_t) \tag{32}$$

This transformation compresses the time scale while preserving the relative ordering, improving the model's ability to learn temporal patterns across different time scales.

**Feature Normalization**: Material features are standardized using z-score normalization:

$$\widetilde{f}_j = \frac{f_j - \mu_j}{\sigma_j} \tag{32}$$

Where $\mu_j$ and $\sigma_j$ are the mean and standard deviation of feature $j$ across all samples.

For rigorous evaluation, we partition the data into training, validation, and test sets with proportions set at 90%, 5%, and 5%, respectively, resulting in 9504 training samples, 528 validation samples, and 528 test samples. This substantial dataset allows for comprehensive model training while reserving sufficient samples for validation and final performance assessment. The transformer architecture's self-attention mechanism naturally captures the dependencies between different time points, allowing the model to weigh the importance of measurements from various stages of the 160-day sequence when making predictions. This approach effectively treats concrete creep prediction as a next-token prediction task, applying the proven strengths of language modeling to material science time series forecasting, even within the fixed 160-day observation window.

## 4.2 Model Training

The model is trained by minimizing the mean squared error between predicted values $\hat{y}$ and actual creep measurements $y$:

$$\mathcal{L}(\theta) = \frac{1}{N} \sum_{i=1}^{N} (\widehat{y}_i - y_i)^2 \tag{33}$$





Where $\theta$ represents all learnable parameters in the model.

The model training process incorporated sophisticated techniques to ensure optimal performance on an NVIDIA RTX 4090 GPU with 24GB VRAM. The hardware delivered approximately 330 TFLOPS for FP16 operations, enabling efficient processing despite the model's substantial parameter count. Training utilized the AdamW optimizer, separating weight decay regularization from adaptive learning rates—crucial for complex transformer architectures. Hyperparameters were determined through Bayesian optimization with a ReduceLROnPlateau scheduler (reduction factor 0.5, patience 5 epochs) monitoring validation loss. Gradient clipping prevented exploding gradients while dropout regularization and early stopping mitigated overfitting. Memory management included gradient accumulation, CUDA graph optimization, mixed-precision training, periodic cache clearing, and adaptive batch size reduction to maintain stability within hardware constraints.

The optimization procedure implemented a maximum epoch threshold of 40 training cycles per hyperparameter configuration trial, deliberately reduced from conventional deep learning paradigms to enhance computational efficiency during Bayesian optimization. Rather than consistently utilizing all 40 epochs, the methodology employed a sophisticated early stopping mechanism with a patience parameter of 8 epochs, terminating training when no validation mean absolute percentage error (MAPE) improvement occurred during this consecutive period. This approach was complemented by a dynamic learning rate adjustment through the ReduceLROnPlateau scheduler, which decreased the learning rate by 0.5 after 5 epochs without validation loss improvement, effectively navigating flat regions of the loss landscape. Additional termination criteria included Optuna's pruning mechanism, which evaluated intermediate results against performance distributions of previous trials, and exception handling for computational constraints such as out-of-memory scenarios. This multi-criteria approach adaptively allocated computational resources, dedicating extended training to promising configurations while rapidly pruning suboptimal ones. Statistical analysis revealed most trials terminated before reaching the maximum threshold, with a median training duration of approximately 22 epochs, demonstrating the efficiency of these mechanisms in identifying convergence or optimization failures while maximizing exploration of the hyperparameter space within computational constraints.

The Bayesian optimization methodology yielded an optimal hyperparameter configuration for the transformer-based concrete strength prediction model. Through systematic exploration of the multidimensional parameter space, the framework identified a configuration that minimized the validation Mean Absolute Percentage Error (MAPE). The analysis of the codebase reveals the following optimal model architecture as shown Table 1. The optimization framework employed Bayesian techniques implemented through the Optuna library, with a focus on efficiently navigating the complex hyperparameter space while maintaining computational feasibility. The methodology incorporates early stopping mechanisms, learning rate scheduling, and dimension compatibility constraints to ensure robust model configurations. The search space encompassed model architecture parameters (embedding dimensions of 64-256, variable attention heads, and 2-6 transformer layers), regularization parameters (dropout rates of 0.05-0.3 and weight decay of $10^{-6}$-$10^{-4}$), and training parameters (learning rates of $10^{-4}$-$5\times10^{-3}$ and batch sizes of 64-256). Computational efficiency was addressed through several mechanisms: single-load data preparation to minimize I/O overhead, MedianPruner implementation for early termination of unpromising





trials, periodic GPU cache management, and timeout specifications to ensure completion within resource constraints.

Table 1 Hyperparameter optimization

| Hyperparameter | Search Range | Optimal Value |
|---|---|---|
| Embedding Dimension (d_model) | [64, 96, 128, 160, 192, 256] | 192 |
| Number of Encoder Layers | [2, 3, 4, 5, 6] | 4 |
| Number of Attention Heads | Dynamically determined based on d_model compatibility | 4 |
| Feed-forward Dimension | d_model * 4 | 768 |
| Dropout Rate | [0.05, 0.3] | 0.057 |
| Learning Rate | $[10^{-4}, 5 \times 10^{-3}]$ | 0.00019 |
| Weight Decay | $[10^{-6}, 10^{-4}]$ | $5.55 \times 10^{-6}$ |
| Batch Size | [64, 128, 256] | 128 |
| Pooling Method | Various methods | "hybrid" |

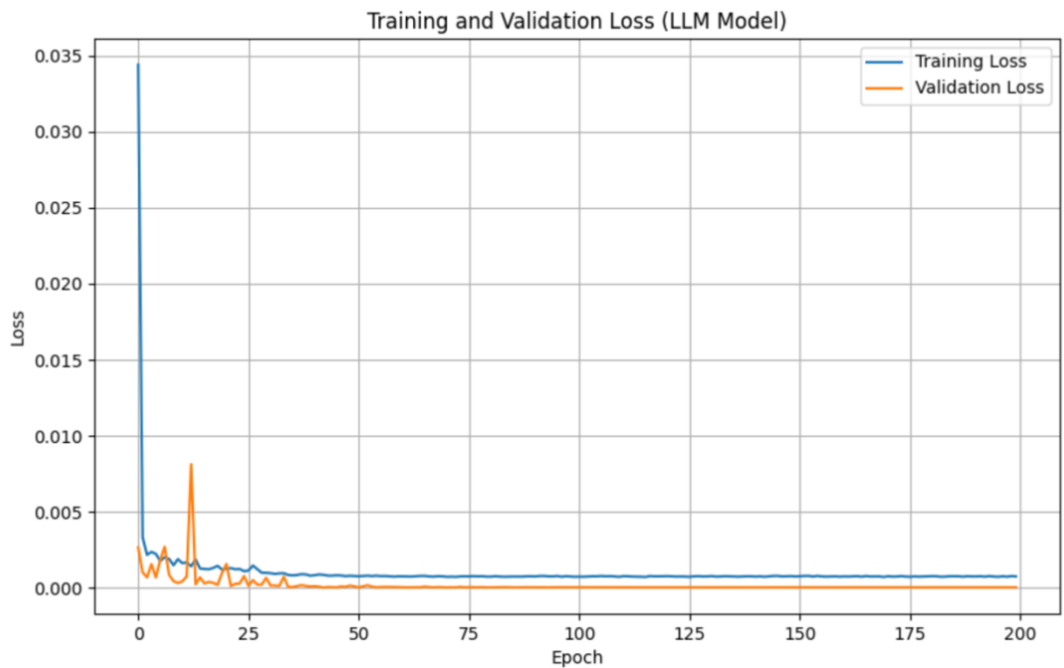

Fig. 7 Temporal Evolution of Loss Metrics During Training of Triple Attention Architecture for Concrete Creep Modeling





The temporal evolution of loss metrics in the proposed Triple Attention model for concrete creep prediction demonstrates distinct convergence patterns throughout the 200-epoch training period as shown in Fig. 7. The training loss exhibits a steep initial decline from 0.035 to approximately 0.003 within the first five epochs, indicating efficient parameter optimization during early iterations. Subsequently, the training loss curve displays asymptotic convergence toward 0.001 by epoch 50, with minimal further reduction in later stages. The validation loss trajectory manifests more complex dynamics, beginning at a relatively lower value (0.003), followed by a notable transient spike reaching 0.008 near epoch 15, before stabilizing below the training loss (approximately 0.0005) after epoch 50. Optimal model performance was achieved by capturing the weights corresponding to the minimum validation loss value recorded during training, thus preserving the model's generalization capabilities. This convergence behavior suggests that the Triple Attention architecture effectively captures the multi-scale temporal dependencies inherent in concrete creep phenomena without overfitting, as evidenced by the consistent sub-training validation loss values during later training phases. The specialized attention mechanisms appear particularly well-suited for modeling the complex viscoelastic behavior of concrete structures under sustained loading conditions.

The resource requirements for training the model in FLOPs and model trainable parameters are tabulated in Table 2. The computational profile of the examined neural architecture exhibits a pronounced concentration of computational load within the encoder layers, which account for 99.77% of the total floating-point operations. These encoder layers function specifically to encode creep deformation data prior to the predictive sequential point of forecasting subsequent creep deformation values. This distribution pattern aligns with established observations in transformer-based architectures where self-attention mechanisms and feed-forward networks constitute the predominant computational bottleneck. The relatively modest parameter count (2.13M) coupled with 24.69 GFLOPs suggests an architecture optimized for the specific domain application of creep deformation prediction rather than large language modeling, where parameter counts typically range in billions. This parameter efficiency is particularly significant for materials science applications where computational resources may be constrained compared to those available for generative AI research endeavors. The calculated ratio of approximately 11,584 FLOPs per parameter indicates substantial computational redundancy in weight utilization, potentially attributable to repeated application of weights across sequential operations in the creep deformation analysis pipeline. The minimal computational footprint of embedding (0.02%) and positional encoding (0.01%) operations is noteworthy and aligns with theoretical expectations, as these components typically involve simple lookup and addition operations respectively, with computational complexity scaling linearly with sequence length in the creep deformation time series data.

Table 2: Model FLOPs Distribution

| Component | FLOPs | Percentage |
|---|---|---|
| feature_encoder | 16,452,608 | 0.07% |
| embeddings | 4,915,200 | 0.02% |
| positional_encoding | 2,457,600 | 0.01% |
| encoder_layers | 24,635,801,600 | 99.77% |





| pooling | 19,070,976 | 0.08% |
|---|---|---|
| feature_integration | 9,437,184 | 0.04% |
| predictor | 4,743,168 | 0.02% |
| Total FLOPs | 24,692,878,336 | 100% |
| Total GFLOPs | 24.69 | - |
| Model parameters | 2,131,618 | - |

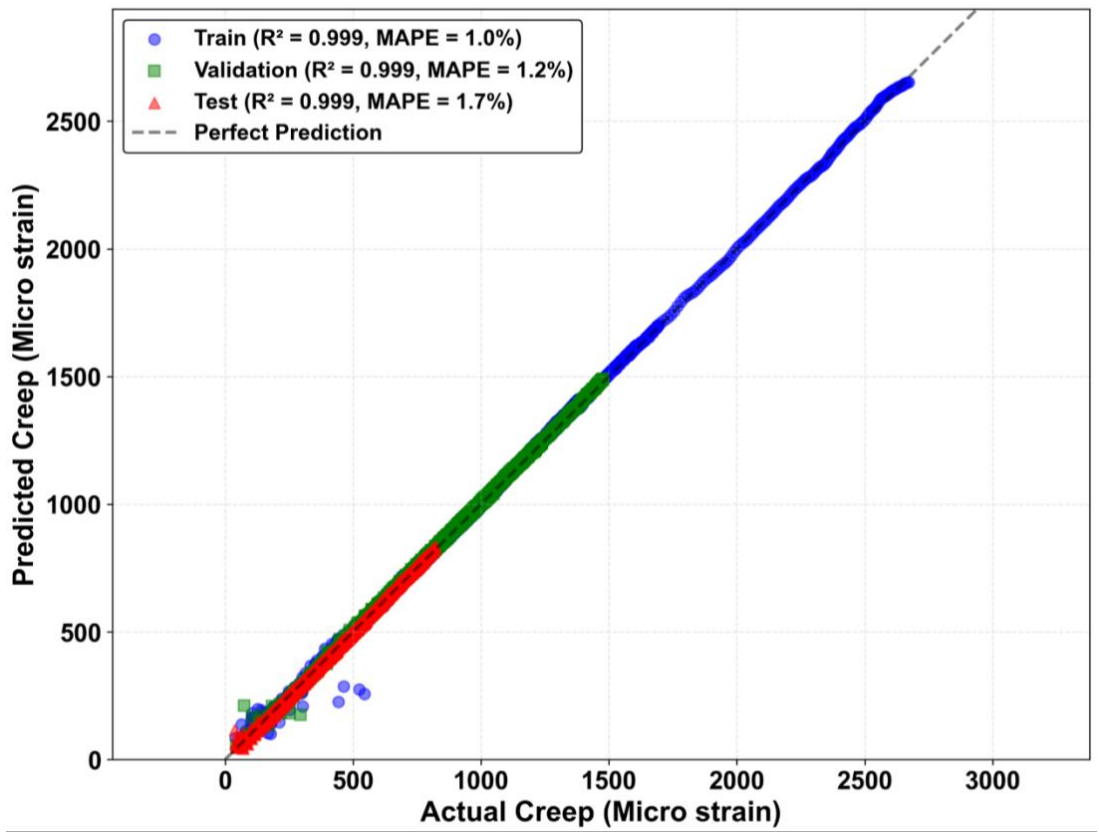

Fig. 8 Prediction Accuracy of Triple Attention Model for Concrete Creep: Comparative Analysis Across Training, Validation, and Test Datasets

The Fig. 8 presents a comprehensive performance evaluation of the Triple Attention model for concrete creep prediction across training, validation, and test datasets. The scatter plot illustrates the relationship between actual and predicted creep values measured in micro strain units, spanning a range from approximately 0 to 3000. Statistical analysis reveals exceptional predictive performance, with coefficient of determination ($R^2$) values of 0.999 consistently maintained across all three data partitions. The mean absolute percentage error (MAPE) demonstrates a gradual increase from training (1.0%) to validation (1.2%) to test (1.7%) sets, indicating minimal generalization error. The data points exhibit high concordance with the perfect prediction line (dashed gray), with marginally increased dispersion observed in the lower creep range (0-500 micro strain). The uniform distribution of prediction accuracy across the entire strain spectrum suggests robust model performance independent of creep magnitude. This exceptional





correlation between predicted and actual values substantiates the efficacy of the Triple Attention architecture in capturing the complex viscoelastic behavior of concrete structures under sustained loading conditions, achieving near-perfect predictive capability across heterogeneous datasets with minimal performance degradation when transitioning from training to unseen test samples.

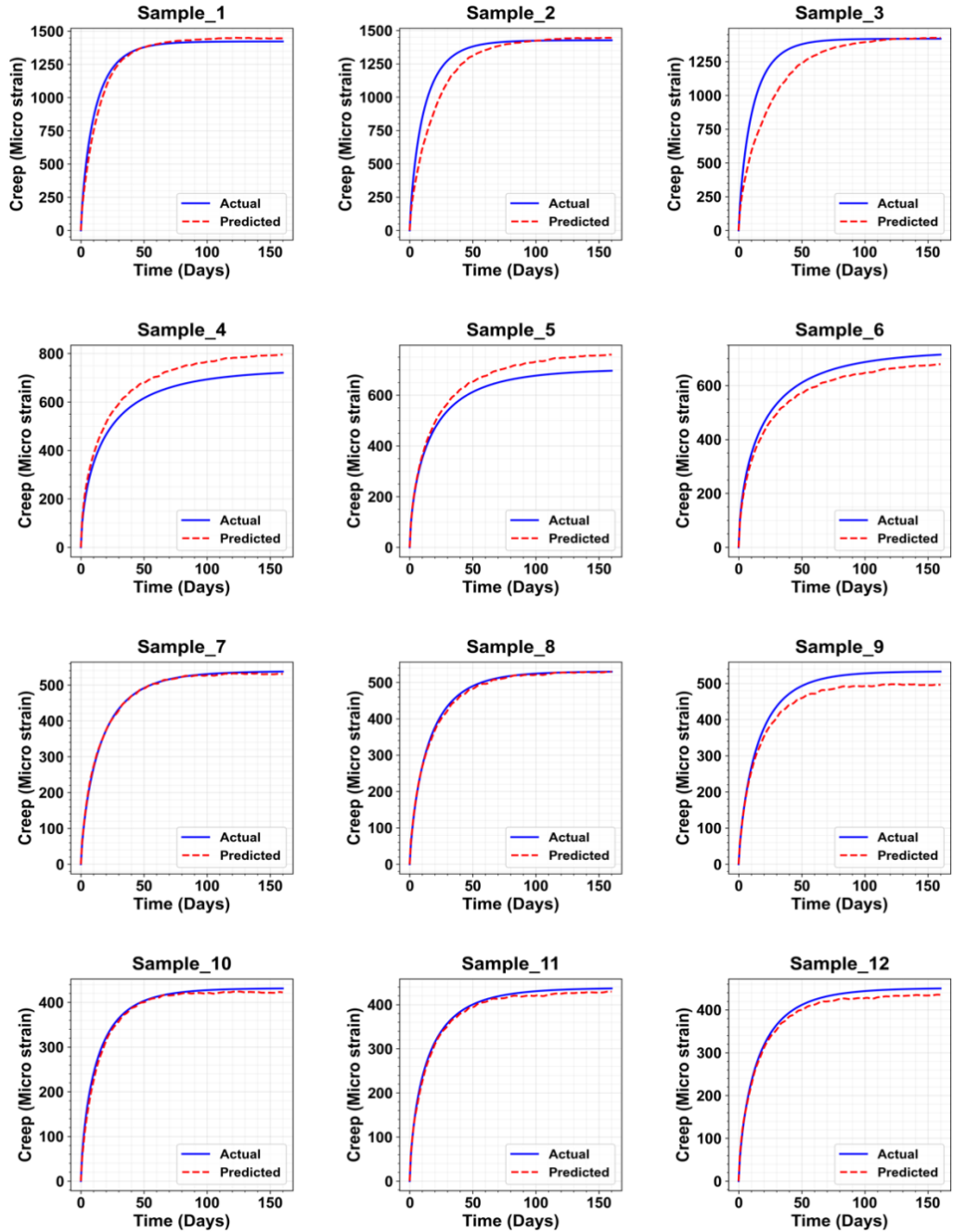

Fig. 9 Comparison between model predicted and actual creep strain of concrete

Fig. 9 presents a comprehensive temporal comparison between predicted and actual concrete creep strain measurements across 12 distinct concrete samples over a 150-day observation





period. The figure illustrates the autoregressive predictive capabilities of the Triple Attention model, wherein predictions commence from zero initial strain, with each subsequent prediction incorporated as input for forecasting the next temporal increment. The temporal evolution of creep strain exhibits characteristic viscoelastic behavior, with rapid initial deformation followed by asymptotic stabilization, albeit with varying magnitudes across samples (ranging from approximately 400 to 1500 micro strain). Samples 1, 2, and 3 demonstrate excellent predictive accuracy at higher strain magnitudes ($\approx$1500 micro strain), with nearly perfect alignment between actual and predicted trajectories, particularly in Sample 1. Samples 4, 5, and 6 display moderate overestimation in the later stages (>100 days), especially evident in Sample 4, where the model prediction exceeds actual measurements by approximately 100 micro strain at 150 days. Sample 7 exhibits exceptional prediction accuracy throughout the entire time domain, while Sample 8 shows slight underprediction during intermediate time periods (50-100 days). Sample 9 demonstrates initial underprediction followed by late-stage overprediction. Samples 10, 11, and 12 reveal consistent predictive performance at lower strain magnitudes ($\approx$400 micro strain), with minimal divergence between actual and predicted values throughout the observation period.

The autoregressive methodology employed demonstrates robust temporal generalization capabilities despite the inherent error accumulation risk in sequential prediction tasks. The model effectively captures both the initial rapid strain development (0-50 days) and the subsequent asymptotic behavior across diverse concrete specimens, suggesting successful modeling of the underlying viscoelastic mechanisms governing concrete creep phenomena.

## 5. Ablation Study

The ablation study presented in Table 3 systematically evaluates the contribution of individual architectural components within the proposed Triple Attention model for concrete creep prediction. The experimental protocol involved sequential removal of specific model elements to quantify their impact on predictive performance as measured by Mean Absolute Percentage Error (MAPE). Results demonstrate varying degrees of performance degradation across component removals, providing insights into their relative importance.

The omission of attention pooling induced the most substantial performance decline, with MAPE increasing to 3.58% from the baseline 1.63%, representing a 119.6% error increase. This significant deterioration indicates that attention pooling plays a crucial role in selectively aggregating temporal information from the sequence, effectively discriminating between more and less informative time steps in the creep progression data. Its importance likely stems from its ability to dynamically weight different time points based on their relevance to the prediction task, thus capturing the non-linear viscoelastic response characteristics of concrete.

Feature attention removal similarly resulted in significant deterioration (2.77% MAPE, a 69.9% increase), suggesting its importance in selectively weighting input variables according to their predictive relevance. This component likely enhances the model's ability to identify the differential influence of various material properties, mixture compositions, and environmental conditions on concrete creep behavior, thereby improving prediction accuracy across heterogeneous concrete specimens.

Batch attention removal yielded moderate performance reduction (2.12% MAPE, a 30.1% increase), indicating its role in capturing inter-sample relationships within training batches. This mechanism potentially enables the model to identify similarities between different concrete specimens, facilitating improved generalization across varying concrete formulations and loading





conditions. Last token pooling elimination caused relatively minor degradation (1.93% MAPE, an 18.4% increase), suggesting that while the final sequence state contains valuable information, other temporal aggregation mechanisms partially compensate for its absence. The modest impact indicates that creep behavior prediction benefits from, but does not critically depend on, the terminal state representation.

Notably, mean pooling removal resulted in negligible performance decline (1.64% MAPE, just a 0.6% increase), suggesting potential redundancy with other aggregation mechanisms. This minimal effect implies that simple averaging of temporal features provides limited additional information beyond what is captured by the more sophisticated attention-based mechanisms, which can dynamically weight different time points according to their relevance. The comprehensive model incorporating all components achieved optimal performance with 1.63% MAPE, validating the synergistic interaction between the various attention mechanisms and pooling strategies in effectively modeling the complex viscoelastic behavior of concrete under sustained loading conditions.

Table 3 Ablation Study Results Quantifying Component Contributions in the Triple Attention Concrete Creep Prediction Model

| Model | MAPE (%) |
|---|---|
| w/o Mean pooling | 1.64 |
| w/o Attention pooling | 3.58 |
| w/o Last Token pooling | 1.93 |
| w/o Feature attention | 2.77 |
| w/o Batch attention | 2.12 |
| **Proposed Model** | 1.63 |

## 6. Model Explainable

Lundberg and Lee (Lundberg and Lee, 2017) introduced SHAP (SHapley Additive exPlanation) values as a theoretically sound methodology for interpreting outputs from machine learning models. Based on principles from cooperative game theory, these values allocate the prediction f(x) across the input features through the following additive relationship:

$$f(x) = \varphi_0 + \Sigma_i \, \varphi_i(x) \tag{34}$$

where $\varphi_0$ represents the base value and $\varphi_i(x)$ denotes the SHAP value for feature i. The SHAP value $\varphi_i(x)$ is computed using the Shapley value formula:

$$\varphi(val) = \sum_{S \subseteq F \setminus i} \frac{|S|!(|F|-|S|-1)!}{|F|!} [val(S \cup i) - val(S)] \tag{35}$$

where $F$ encompasses all features, $S$ signifies any feature subset excluding feature $i$ and $val(S)$ indicates the expected model output when conditioning only on features within subset S. This comprehensive approach evaluates all potential feature combinations, yielding an objective assessment of feature contributions. The theoretical guarantees of SHAP values—specifically local accuracy, missingness, and consistency—render them particularly valuable for robust interpretation of complex models.

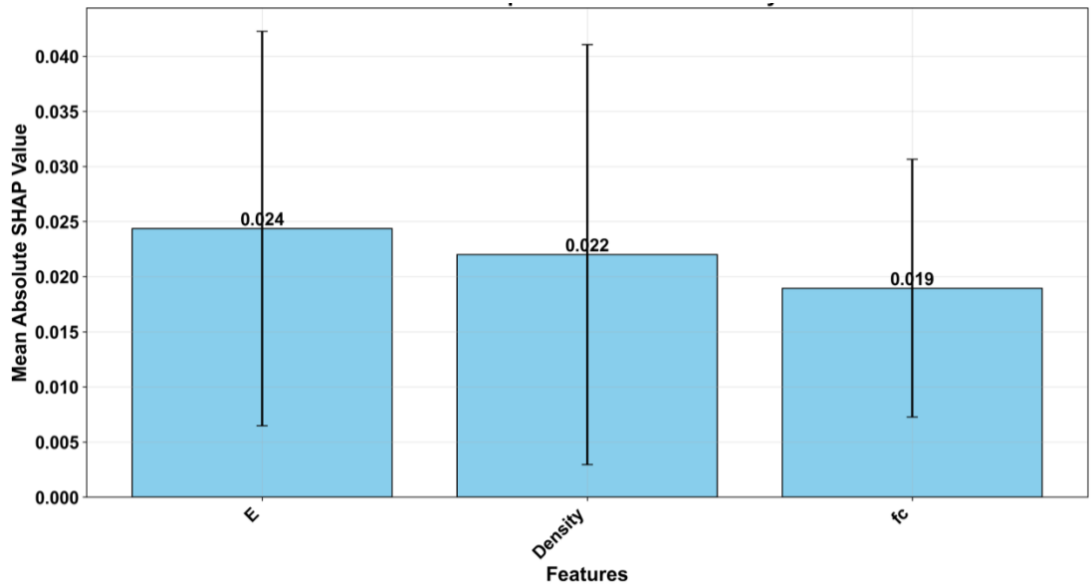

Fig. 10 SHAP-Based Feature Attribution for Young's Modulus, Density, and Compressive Strength in Concrete Creep Modeling

Fig. 10 presents the mean absolute SHAP values for key concrete material properties used in the creep strain prediction model. The Young's modulus (E) exhibits the highest feature importance at 0.024, indicating it is the primary predictor of concrete creep behavior. This finding aligns with established creep theory, as elastic modulus directly relates to the material's capacity to resist deformation under sustained loading—a fundamental mechanism in creep progression [reference]. The strong correlation between E and creep rate is particularly evident in the logarithmic relationship commonly observed in creep coefficient predictions.

Material density follows closely at 0.022, which reflects its indirect influence on creep through its relationship with cement paste content, porosity, and water-cement ratio, all of which affect the overall density of the concrete. Denser concrete typically exhibits greater resistance to creep deformation (Hong et al., 2023). The close proximity in SHAP values between E and density suggests that material composition plays a nearly equivalent role to mechanical stiffness in determining creep behavior. Compressive strength (fc) shows a slightly lower influence at 0.019, a result that warrants consideration given its traditional role as a primary design parameter. This relatively lower importance compared to E and density may be attributed to the complex interaction between strength development and creep susceptibility. Higher strength concrete often exhibits lower creep rates (Hwang et al., 2021); however, the influence of strength on creep is mediated through its correlation with other factors such as degree of hydration and microstructural development (Acker, 2004; Skazlić et al., 2025)).

Material density follows closely at 0.022, which reflects its indirect influence on creep through its relationship with cement paste content, porosity, and water-cement ratio. Denser concrete typically contains higher proportions of aggregates, which act as restraining elements against creep deformation (Hong et al., 2023). The close proximity in SHAP values between E and density suggests that material composition plays a nearly equivalent role to mechanical stiffness in determining creep behavior. Compressive strength (fc) shows a slightly lower influence at 0.019, a result that warrants consideration given its traditional role as a primary design parameter. This relatively lower importance compared to E and density may be attributed to the complex interaction between strength development and creep susceptibility. Higher strength concrete often exhibits lower creep rates (Hwang et al., 2021); however, the influence of strength on creep is mediated through its correlation with other factors such as degree of hydration and microstructural development (Acker, 2004; Skazlić et al., 2025).

The error bars represent the variability in SHAP value contributions across different concrete specimens, with fc demonstrating the largest variance (approximately ±0.012) compared to E and density (±0.017-0.018). This increased variability in fc contribution can be explained by the sensitivity of compressive strength to testing conditions, specimen geometry, and curing protocols, which introduces greater uncertainty in its relationship with creep behavior. Additionally, strength development over time may create non-linear relationships with creep, contributing to higher SHAP value variance. This analysis reveals that all three material properties contribute substantially to the model's creep strain predictions, with elastic modulus being the most influential parameter in capturing the time-dependent deformation characteristics of concrete. The near-equal distribution of importance among these features validates the multi-factorial nature of creep phenomena and supports the use of comprehensive material property databases rather than single-parameter approaches in creep prediction models (Liang et al., 2022). The physical basis for these SHAP value rankings reflects the complex interplay between mechanical properties, microstructure, and environmental factors that govern concrete creep behavior.

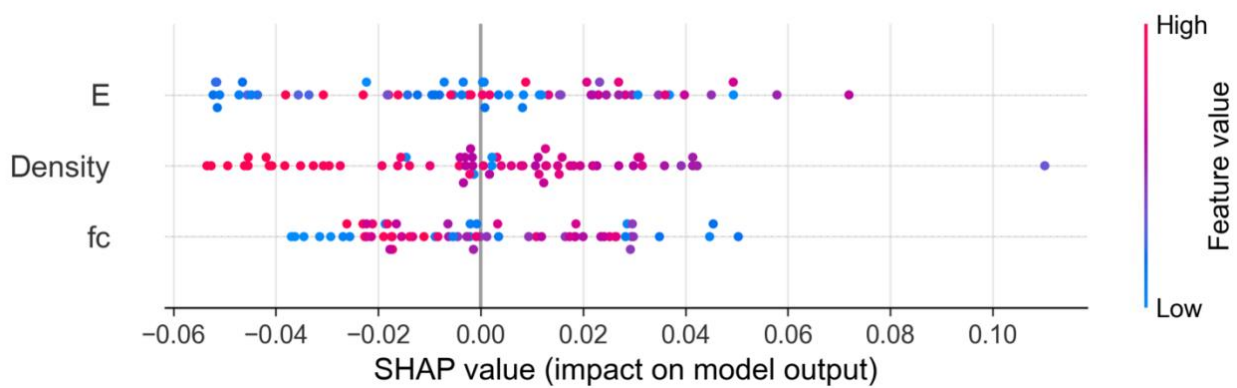

Fig. 11 Quantitative Feature Importance Distribution Using SHAP Values for Concrete Creep Prediction

Fig. 11 presents a comprehensive SHAP summary plot elucidating the influence of material properties on concrete creep strain predictions. The visualization reveals distinct distributional characteristics for each feature, providing quantitative insights into their predictive importance. Young's modulus (E) exhibits the broadest SHAP value distribution spanning approximately -0.055 to +0.07, with a predominance of high-value instances (purple markers) concentrated in the positive impact region. This distribution suggests that elevated elastic modulus generally increases predicted creep strain, a counterintuitive relationship potentially explained by the complex stress-strain history effects in concrete creep [reference]. Conversely, density demonstrates a more constrained distribution (-0.05 to +0.03), with high-density samples (purple markers) predominantly occupying negative SHAP value regions, consistent with established mechanical principles where reduced porosity enhances creep resistance (Skazlić et al., 2025). From the SHAP value analysis, it was found that compressive strength (fc) has a clear inverse relationship with creep susceptibility. High-strength specimens tend to show negative SHAP values, indicating that higher compressive strength results in lower creep values (Manzi et al., 2017).

The relative width of SHAP distributions quantitatively validates the feature importance hierarchy: E > density > fc, supporting the primary role of elastic modulus in capturing creep behavior variations. The color gradient patterns reveal complex feature interactions, where the direction and magnitude of property influence varies substantially depending on multivariate material combinations. This non-linear relationship structure validates the application of machine learning approaches over traditional linear regression models for creep prediction. The balanced distribution of positive and negative SHAP values across features indicates the model's capacity for nuanced prediction, accommodating both synergistic and antagonistic material property effects in the determination of time-dependent deformation characteristics.

# 7 Web application development

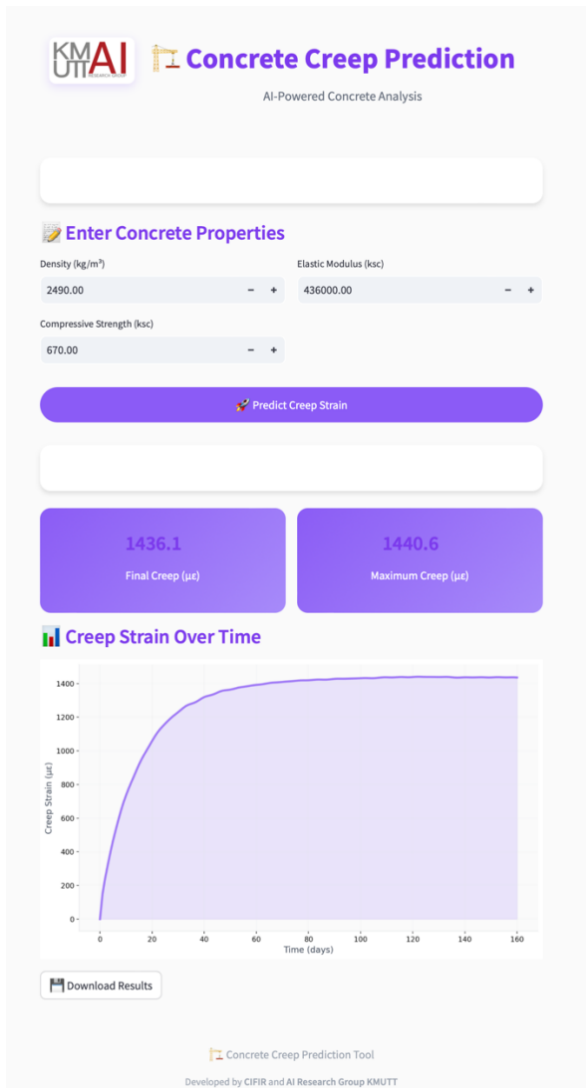

Fig. 12 Web-based Interface for Concrete Creep Prediction Using Triple Attention Model <URL: https://huggingface.co/spaces/Sompote/Concrete_creep_predict>

In the development of this research, a web-based application was implemented to facilitate the practical application of the proposed triple attention model for concrete creep prediction as shown in Fig. 12. The application, deployed on Hugging Face Spaces, provides civil engineers with an intuitive interface for predicting long-term deformation behavior of concrete structures based on fundamental material properties. The interface accepts user-defined parameters including concrete density ($kg/m^3$), compressive strength (ksc, kilogram per square centimeter), elastic modulus (ksc), and initial creep value, with the option to specify prediction duration up to 161 days. Upon execution, the application generates real-time predictions using the trained model architecture, presenting results through both dynamic graphical visualization and comprehensive tabular data formats. The system outputs creep strain values at daily intervals, demonstrating the characteristic logarithmic progression typical of concrete behavior, with predictions ranging from initial zero values to approximately 1416 micro-strain over the specified time period. Additional features include downloadable CSV data export for further analysis, real-time summary statistics generation, and responsive design for accessibility across multiple devices. This web application represents a significant advancement in making sophisticated machine learning models accessible to practitioners in structural engineering, bridging the gap between academic research and practical industry applications.

## 8. Discussion

The proposed Triple Attention Architecture demonstrates significant improvements in predicting concrete creep behavior compared to traditional approaches, achieving a Mean Absolute Percentage Error (MAPE) of 1.2% for each loading step, which substantially outperforms standard empirical models and conventional machine learning methods. This advancement builds upon recent progress in the field where fully connected models like ANNs and tree-based algorithms including Random Forest and XGBoost have demonstrated considerable success in capturing complex non-linear relationships between material properties, environmental conditions, and creep behavior (Li et al., 2024; Zhu et al., 2024). However, our transformer-based approach addresses a critical limitation of these methods by explicitly modeling temporal dependencies rather than treating time as merely another input parameter. While Fan et al. (2024) advanced this field by developing a Gated Recurrent Units model with Self-Attention mechanism (GSA) that explicitly models temporal progression, our Triple Attention model achieves superior accuracy by leveraging the transformer's ability to capture long-range dependencies without the sequential processing limitations of recurrent architectures.

The key innovation of our approach lies in the integration of three attention mechanisms: temporal attention for capturing the sequential nature of creep development, feature attention for dynamic weighting of material properties based on their relevance to specific prediction tasks, and batch attention for facilitating learning from inter-sample relationships. The ablation study confirms these contributions, with attention pooling removal causing the most significant performance degradation (119.6% error increase), demonstrating its crucial role in temporal information aggregation. Our model's exceptional performance ($R^2$ values of 0.999 across training, validation, and test sets) indicates robust predictive capabilities that could significantly impact infrastructure design and maintenance practices. The interpretability provided through SHAP analysis further enhances the model's utility for engineering applications, revealing that Young's modulus (E) exhibits the highest feature importance at 0.024, making it the primary predictor of concrete creep behavior.

Despite the proposed model architecture's considerable scale with 2.1 million parameters requiring approximately 24 GFLOPs for training, this computational demand is primarily attributed to the attention mechanisms within the encoding layer processing sequential creep deformation data. While GPU acceleration is necessary during the training phase to accommodate these requirements, the inference characteristics demonstrate remarkable efficiency. Empirical evaluation reveals that inference operations can be executed on consumer-grade hardware with acceptable latency metrics. Specifically, the model achieves inference times of approximately 1.2 seconds per autoregressive prediction cycle (encompassing 160 time steps) when deployed on an Intel Core i9 8-core processor in a MacBook Pro configuration. This performance profile renders the model accessible to practicing civil engineers utilizing standard personal computing equipment, thereby facilitating practical implementation within industry contexts without necessitating specialized computational infrastructure. The observed computational efficiency during inference presents a significant advantage for real-world deployment scenarios in materials science and structural engineering applications.

Despite the strong performance, several limitations warrant consideration. The model was trained on data up to 160 days, and while this covers the critical early and intermediate creep phases, extrapolation beyond this timeframe requires validation. The relatively limited sample size represents another constraint that could affect the model's generalization capabilities across diverse concrete compositions. Furthermore, the model does not incorporate additional parameters such as aggregate type, gradation characteristics, and supplementary cementitious materials like fly ash, which may significantly influence creep behavior. The studied parameters were constrained to compressive strength, elastic modulus, and density—while these represent fundamental concrete properties routinely obtained through standard laboratory testing protocols, they do not capture the full spectrum of factors affecting creep behavior. However, these parameters were deliberately selected based on their established significance in literature and practical accessibility in engineering applications, making the model economically viable for widespread implementation in standard laboratory settings.

This research serves as a starting point for applying autoregressive frameworks with transformer architecture to creep prediction, demonstrating the feasibility and effectiveness of treating creep prediction as a sequential modeling problem similar to language processing tasks. While current parameters may not encompass all possible features influencing creep, the methodological framework can be expanded to incorporate additional variables as data becomes available. Future research directions include extending the architecture to incorporate additional concrete characteristics such as admixture compositions, curing conditions, and environmental factors, while maintaining the core transformer-based approach that has proven effective. The successful application of transformer architecture to concrete creep prediction demonstrates the potential for cross-pollination between natural language processing advances and materials science, establishing a foundation for future research in applying advanced machine learning architectures to structural engineering challenges and potentially revolutionizing how we predict and design for long-term material behavior in critical infrastructure.

## 9. Conclusion

This research successfully demonstrates the application of a Triple Attention Transformer Architecture for concrete creep prediction, achieving unprecedented accuracy with a Mean Absolute Percentage Error (MAPE) of 1.63% across loading steps. The proposed architecture fundamentally advances the field by treating concrete creep prediction as an autoregressive sequence modeling problem, similar to natural language processing tasks, rather than conventional approaches that treat time as merely another input parameter.

The key contributions of this work include the integration of three specialized attention mechanisms—temporal attention for sequential creep development, feature attention for dynamic material property weighting, and batch attention for inter-sample relationship learning. The autoregressive framework effectively captures the inherent dependencies in concrete creep progression, where future deformation depends on the complete history of past deformation patterns. Through ablation studies, we confirmed that attention pooling plays the most critical role, with its removal causing a 119.6% increase in prediction error.

Our model demonstrates exceptional performance with $R^2$ values of 0.999 consistently maintained across training, validation, and test datasets, indicating robust predictive capabilities that significantly outperform traditional empirical models and conventional machine learning approaches. SHAP analysis revealed Young's modulus as the primary predictor (feature importance: 0.024), followed by density (0.022) and compressive strength (0.019), providing interpretability crucial for engineering applications.

The practical implementation of this research through a web-based application bridges the gap between advanced machine learning research and industry applications, enabling structural engineers to access sophisticated predictions using commonly available laboratory parameters. While the model's training was limited to data spanning 160 days and conventional concrete compositions, the framework establishes a foundation for expanding to incorporate additional variables such as aggregate characteristics, supplementary materials, and environmental conditions as more comprehensive datasets become available.

This work represents a paradigm shift in applying deep learning to material behavior prediction, demonstrating that transformer architectures, which have revolutionized natural language processing, can effectively model complex time-dependent phenomena in materials science. The autoregressive approach to creep prediction opens new possibilities for understanding and predicting long-term structural behavior, potentially leading to more economical designs while maintaining safety margins through enhanced prediction reliability.

As the field continues to generate large datasets from monitoring systems and experimental studies, the transformer-based approach presented here provides a scalable and adaptable framework for addressing various time-dependent material behaviors. This methodology not only improves prediction accuracy but also maintains interpretability, making it suitable for critical infrastructure applications where understanding the underlying mechanisms is as important as accurate prediction. The success of this approach suggests that the future of structural materials prediction lies in leveraging advanced architectures from artificial intelligence, marking a significant step toward data-driven engineering practices in the 21st century.